%% file: main.tex
\documentclass[11pt]{article}

\usepackage[T1]{fontenc}
\usepackage{microtype}
\usepackage{graphicx}
\usepackage{booktabs}
\usepackage{placeins}
\usepackage{flafter}
\usepackage{multirow}
\usepackage{amsmath}
\usepackage{amssymb}
\usepackage{xcolor}
\usepackage{enumitem}
\usepackage[ruled,vlined,linesnumbered]{algorithm2e}
\usepackage[normalem]{ulem}

\PassOptionsToPackage{hyperfootnotes=false}{hyperref}
\input{style/venue}

\graphicspath{{figures/}}

\newif\ifdraftnotes
\draftnotesfalse

\setlist[enumerate,itemize]{itemsep=0pt, leftmargin=*, after=\leavevmode}
\setlist[enumerate,1]{label=(\arabic*)}

\title{VistaTree: Question Answering over Long Multimodal Documents\\
with Image-Derived Summary Trees}
\author{
  Yexiang Cheng\textsuperscript{}\quad
  Xiaochen Zheng\textsuperscript{}\quad
  David Tokar\textsuperscript{}\quad  Tatyana Y. Doktorova\textsuperscript{} \\
  \normalfont\textsuperscript{}Translational Safety Assessment and Computational Toxicology, Roche, Basel, Switzerland \\
  \normalfont\texttt{yexiang.cheng@roche.com}
}
\date{}

\begin{document}
\raggedbottom
\maketitle

\begin{abstract}
\input{sections/en/00_abstract}

\end{abstract}

\input{sections/en/01_introduction}
\input{sections/en/02_related_work}
\input{sections/en/03_method}
\input{sections/en/04_experimental_setup}
\input{sections/en/05_results}
\FloatBarrier
\input{sections/en/06_conclusion}

\FloatBarrier
\section*{Limitations}
\label{sec:limitations}
\input{sections/en/07_limitations}

\bibliography{refs}

\appendix
\input{sections/en/08_appendix}

\end{document}

%% file: style/venue.tex
\ifdefined\VistaTreeReview
  \usepackage[review]{style/acl}
\else
  \ifdefined\VistaTreeFinal
    \usepackage[final]{style/acl}
  \else
    \usepackage[preprint]{style/acl}
  \fi
\fi
\usepackage{times}
\usepackage{latexsym}

\definecolor{ACLLinkBlue}{HTML}{000099}
\hypersetup{colorlinks=true,citecolor=ACLLinkBlue,
  linkcolor=ACLLinkBlue,urlcolor=ACLLinkBlue}

%% file: sections/en/00_abstract.tex
Answering questions over long multimodal documents requires locating relevant evidence across pages containing text, tables, and figures. When the document exceeds the model's context window, selective retrieval becomes necessary. We introduce VistaTree, a document question-answering agent that builds a summary tree directly from rendered page images. VistaTree segments documents into contiguous page spans, recursively groups them into a topic hierarchy, and refines sibling summaries to distinguish neighboring sections. At query time, the agent uses node summaries to select relevant nodes and reads the corresponding pages, using hierarchical navigation when the full outline exceeds the context budget. Experiments on MMLongBench-Doc and LongDocURL evaluate answer quality and retrieval cost using a locally hosted vision--language model, and quantify the performance gap between global and hierarchical access to the index.

%% file: sections/en/01_introduction.tex
\section{Introduction}
\label{sec:intro}

Question answering over long multimodal documents requires finding a small amount of evidence within a much larger source. Clinical study reports, for example, can approach a thousand pages and contain sensitive patient information, which may restrict their processing with externally hosted models. Locally hosted vision--language models offer an alternative, but processing all page images at once can exceed their input budget. Longer contexts also do not guarantee effective use of the available evidence~\citep{liu2024lost}. These constraints motivate selective access to evidence across pages and modalities.

Retrieval-augmented generation (RAG) addresses the input constraint by providing selected passages to a language model~\citep{lewis2020rag}. Dense retrievers typically rank passages through similarity in a shared embedding space~\citep{karpukhin2020dpr}. In long technical documents, repeated topics and terminology can produce many competing candidates. Passages discussing these topics may occupy nearby regions of the embedding space even when they refer to different entities, reporting periods, or measurements. Similarity can therefore capture topical overlap while leaving the distinctions that determine relevance to a question unresolved. Instruction-following retrieval evaluations likewise show that retrievers can struggle with detailed relevance conditions~\citep{weller2025followir}. Increasing the number of retrieved chunks broadens coverage while consuming more of the model's limited context with potentially redundant or unhelpful material. More retrieved text does not necessarily produce better answers~\citep{liu2024lost, jin2025longcontext}. This motivates a search process that compares candidate regions against the question and selects which content merits closer reading.

Beyond selecting relevant passages, text-based retrieval faces the challenge of preserving information from multimodal pages. Document parsers such as MinerU, Docling, and olmOCR recover text and document structure from pages~\citep{wang2024mineru, livathinos2025docling, poznanski2025olmocr}. However, the text supplied to retrieval may omit visual information carried by charts, figures, and spatial relationships on the page. Fixed-size chunks can also separate a result from the context needed to interpret it. These limitations motivate deriving a searchable representation directly from rendered pages and retaining access to the original visual evidence.

We introduce VistaTree, an agent that builds and searches a summary tree directly from rendered document pages. A vision--language model identifies consecutive spans of related content, summarizes them, and recursively groups their summaries into broader topics. Each node remains linked to its underlying page range. At query time, the agent uses the node descriptions to select relevant spans and reads their page images to gather evidence. It can inspect additional spans before producing an answer.

To distinguish spans with overlapping topics, the agent rewrites each node's title and summary using its content together with its siblings' summaries. The tree structure and page ranges remain fixed. Flat selection presents all leaf summaries together for comparison across the document. For longer documents whose complete outlines may exceed the context budget, hierarchical navigation reveals broader topics first and provides child descriptions as the agent descends.

We evaluate VistaTree on MMLongBench-Doc~\citep{ma2024mmlongbench} and LongDocURL~\citep{longdocurl} using the same locally hosted model. These benchmarks allow both flat selection and hierarchical navigation to operate, enabling us to quantify the answer-quality gap between global and progressive access to the index. Flat selection outperforms DocAgent on both benchmarks and Chunk-RAG and RAPTOR on LongDocURL. Summary rewriting improves answer quality for both navigation modes. Hierarchical navigation uses a smaller initial outline, with total query cost varying across datasets. Increasing the passage budget of the retrieval baselines improves evidence coverage more than answer quality, reinforcing the importance of selecting useful context.

Our contributions are:
\begin{enumerate}
\item We introduce VistaTree, a document question-answering agent that constructs a hierarchical summary index directly from rendered page images using a locally hosted vision--language model.
\item We develop summary rewriting that uses sibling context to distinguish related document spans, together with flat and hierarchical search procedures that guide the agent to page images for evidence gathering.

\item We evaluate VistaTree on two public benchmarks, demonstrating the benefits of summary rewriting and global access to document summaries, and characterizing the trade-offs between answer quality and context usage.

\end{enumerate}

%% file: sections/en/02_related_work.tex
\section{Related Work}
\label{sec:related}

\paragraph{Retrieval-augmented generation (RAG).}
Retrieval-augmented generation supplies a language model with information selected from an external collection~\citep{lewis2020rag}. Dense retrieval commonly performs this selection by ranking passages according to their embedding similarity to a question~\citep{karpukhin2020dpr}. Answer quality therefore depends on how well the model uses the retrieved passages. \citet{liu2024lost} show that language models can struggle to use relevant evidence within long contexts. In retrieval-augmented settings, \citet{jin2025longcontext} further find that increasing the number of retrieved passages can reduce answer quality by introducing hard negatives. Their findings highlight how related but unhelpful passages can interfere with the model's use of evidence.

\paragraph{Visual document understanding.}
Document understanding also requires interpreting how information is presented on a page, including the arrangement of text, tables, and figures. 
Donut~\citep{kim2022donut} and Pix2Struct~\citep{lee2023pix2struct} demonstrated that document representations can be learned directly from pixels without a separate OCR stage. For text-based retrieval, document parsers convert pages into structured text such as Markdown or JSON. Modular toolkits such as MinerU~\citep{wang2024mineru} and Docling~\citep{auer2024docling, livathinos2025docling} recover text, layout, and document structure. End-to-end models such as Nougat~\citep{blecher2024nougat} and olmOCR~\citep{poznanski2025olmocr, poznanski2025olmocr2, poznanski2026olmocr} transcribe page images with a single vision encoder-decoder or fine-tuned vision-language model. These outputs can be linearized and chunked for text-based retrieval; information carried by figures and page layout may not survive this step, and parsing errors can propagate to retrieval. General-purpose vision-language models such as Qwen3-VL~\citep{qwen3vl} offer an alternative by answering questions directly from page images. 
VistaTree follows this route: a single vision-language model reads page images both to build the index and to answer questions. Its node descriptions guide the selection of page spans, while evidence is read from the rendered pages.
\paragraph{Hierarchical document retrieval.}
RAPTOR~\citep{sarthi2024raptor} recursively clusters and summarizes text chunks into a multilevel index that connects related content across a document. It retrieves passages and summaries by embedding similarity through either tree traversal or collapsed-tree search. PageIndex~\citep{pageindex} organizes document sections and their descriptions into a hierarchical index that a language model navigates through reasoning-based retrieval. It uses extracted document text to generate node descriptions. VistaTree builds its hierarchy and summaries directly from rendered pages, allowing visual content to inform the index as well as the final reading stage. We further rewrite each summary in the context of its siblings to distinguish related regions while retaining the same tree and page ranges.

\paragraph{Agentic document question answering.}
Agentic methods give a language model tools for choosing which document content to inspect. PDFTriage~\citep{saadfalcon2024pdftriage} combines document-structure metadata with retrieval functions, allowing the model to request context by structure or content. DocAgent~\citep{sun2025docagent} constructs an outline from parsed document content. Its actor navigates this outline and uses tools to retrieve individual elements for multimodal reading. In VistaTree, index construction follows an agent loop: the model reads pages sequentially and decides whether to continue the current span or close it and create a node. At query time, the agent navigates the resulting summaries, reads selected page images, and decides whether to gather more evidence or answer.

%% file: sections/en/03_method.tex
\section{Method}
\label{sec:method}

\begin{figure*}[t]
\centering
\includegraphics[width=\textwidth]{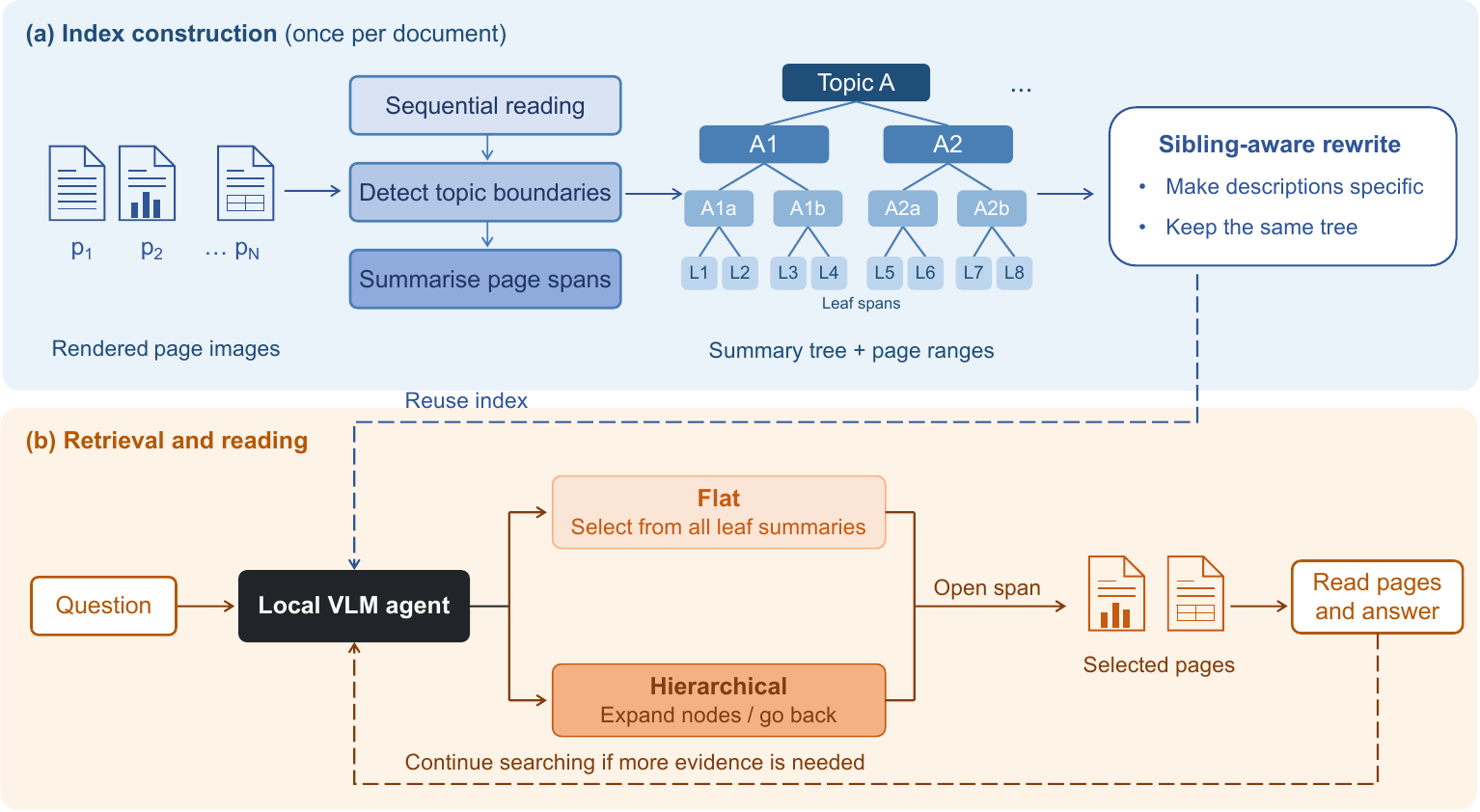}
\caption{Overview of VistaTree. Both stages use the same locally hosted VLM. (a) The model reads pages sequentially to build a tree of consecutive page spans. Summary rewriting refines the node titles and summaries without changing the tree structure. (b) Given a question, the model uses the titles and summaries to select spans, reads their page images, and continues searching when more evidence is needed. Flat selection exposes all leaf summaries, while hierarchical navigation reveals children on demand and supports backtracking. Nodes and page contents are schematic.}
\label{fig:overview}
\end{figure*}

VistaTree uses a vision--language model within an agent loop to construct and search a summary tree. Figure~\ref{fig:overview} shows the two phases of VistaTree: summary-tree construction and question answering. Given rendered pages $D=(p_1,\ldots,p_N)$, VistaTree constructs a tree whose nodes contain a title, a summary, and a page range. The title and summary together form the node description used for navigation. Each leaf represents a span of consecutive pages covering related content, while internal nodes summarize groups of child nodes. At query time, the agent uses the node descriptions to select relevant spans and opens their page images to gather evidence.

\subsection{Constructing the summary tree}
\label{sec:method:tree}
The vision--language model first divides the rendered document into consecutive page spans to create leaf nodes. It then recursively groups nodes and summarizes their content to build a hierarchy whose depth depends on the resulting groups.

To identify the spans, the model reads the document in an agent loop. At each step, it receives the next page image together with a short running description of the current span. It updates this description as long as new pages continue the current section or topic, and closes the span when it identifies a transition. When a span closes, the model creates a leaf node with its page range and generates the node's title and summary from the accumulated description and the span's page images. Reading then continues with a new span until the document has been processed. If a new span begins partway through a page, adjacent spans can share that page.

To build the next level, the model examines the leaf summaries in document order, deciding which consecutive nodes belong together and where a new group should begin. It creates a parent node for each group, generating its title and summary from the children's summaries and retaining their combined page range. The model then processes the parent summaries in the same way to construct further levels. This process repeats until the number of top-level nodes falls below a configured threshold or a safety limit is reached.

The completed tree can be reused across questions. Sampling rules and construction settings are given in Appendix~\ref{app:impl}.

\subsection{Summary rewriting}
\label{sec:method:contrastive}

Neighboring spans often cover similar topics, making them difficult to distinguish from their summaries alone. For example, two spans in a clinical study report may both be summarized as ``liver function test results,'' even though they report outcomes for different dose groups. During hierarchical navigation, the agent chooses among sibling nodes that share a parent. Using their summaries as context when rewriting each node helps make these distinctions explicit.

During a second summarization pass, the agent rewrites each target node's title and summary using its siblings' summaries as context to make the nodes more distinguishable during retrieval. It combines this context with the target's page images for a leaf node, or with its children's summaries for an internal node. If a sibling's summary provides insufficient detail, the agent can inspect that sibling's page images before completing the rewrite.

The agent uses a frozen snapshot of the original summaries for every rewrite, making the results independent of processing order. It applies the updated titles and summaries together after the pass, retaining the tree structure and page ranges. We refer to the original summaries as \textbf{v0} and the rewritten summaries as \textbf{v1}.

\subsection{Searching and reading}
\label{sec:method:nav}

Given a question, the agent uses the visible node descriptions to select relevant page spans and reads their images to gather evidence. After reading a span, the agent can use what it has found to select another span or produce an answer. Its main action, \texttt{expand\_node}, returns child summaries for an internal node and page images for a leaf. Hierarchical navigation additionally supports \texttt{go\_back} to revisit an ancestor and explore another branch. As the search progresses, images from earlier turns are removed from the conversation to leave room for newly retrieved pages.

The nodes presented to the agent, together with their descriptions and page ranges, form an outline. The outline, conversation history, and selected page images share the model's context window. Flat selection presents all leaf descriptions together, allowing comparisons across the document. For substantially longer documents, such as clinical study reports that can exceed a thousand pages, the complete outline may no longer fit within the available context. Hierarchical navigation addresses this constraint by exposing the tree progressively. We compare both configurations on the public benchmarks to assess how hierarchical navigation affects answer quality and total input tokens.

\paragraph{Flat selection.}
The initial prompt lists every leaf with its title, summary, page range, and its parent's title. The agent can compare spans across the document and open a leaf directly. The parent title supplies broader context for interpreting each leaf description. This provides global visibility, but the initial outline grows with the number of leaves.

\paragraph{Hierarchical navigation.}
The initial prompt lists only the top-level nodes. The agent uses \texttt{expand\_node} to inspect their children and descend toward relevant page spans, and \texttt{go\_back} to return to an ancestor and explore another branch. By revealing the tree progressively, hierarchical navigation reduces the size of the initial outline compared with flat selection. However, reaching relevant page spans requires additional routing decisions based on the summaries visible at each step.

%% file: sections/en/04_experimental_setup.tex
\section{Experimental Setup}
\label{sec:setup}

\paragraph{Benchmarks.}
We evaluate VistaTree on two benchmarks for question answering over long multimodal documents. Table~\ref{tab:datasets} summarizes the datasets used in our experiments.

MMLongBench-Doc \citep{ma2024mmlongbench} covers seven document categories, including academic papers, financial reports, brochures, and guidelines. Its questions draw on text, tables, charts, images, and layout, with evidence located on a single page or spread across multiple pages. It also includes unanswerable questions, testing whether a system can recognize when the document does not supply an answer.

LongDocURL \citep{longdocurl} organizes its questions into document understanding, numerical reasoning, and cross-element locating, with 20 subtasks spanning text, tables, figures, and layout. Questions can require combining evidence across pages or identifying relationships between document elements. Its longer median document length complements MMLongBench-Doc's broader length range, allowing us to examine how document length affects navigation and context cost.
\input{tables/datasets}

\paragraph{Models.}
We use Qwen3-VL-8B-Instruct \citep{qwen3vl} for both tree construction and question answering, balancing tree-construction speed with the visual capability needed to extract useful document information. Query experiments use a locally hosted deployment with a 98,304-token context limit and temperature zero. We additionally evaluate Claude Opus 4.8 for question answering using full documents, evidence windows, or summary trees constructed by the local model. Appendix~\ref{app:impl} gives further configuration details.

\paragraph{Retrieval baselines.}
We compare VistaTree with passage retrieval, summary-based retrieval, and agentic document search. Chunk-RAG retrieves passages from extracted text by embedding similarity. RAPTOR \citep{sarthi2024raptor} builds and searches a multilevel index of text chunks and recursively generated summaries. DocAgent \citep{sun2025docagent} retrieves document content through an interactive interface over a structured outline. Its agent uses tools to read extracted text and inspect page images and extracted images. All three use the same local model for question answering. We report Chunk-RAG and RAPTOR on LongDocURL only. Baseline configurations are given in Appendix~\ref{app:baselines}.

\paragraph{Evidence-conditioned references.}
A gold-page oracle supplies annotated evidence images directly to the model. On MMLongBench-Doc, it is evaluated on answerable questions with valid evidence-page annotations and serves as an empirical upper-bound reference under ideal evidence selection. A window reference supplies a contiguous span positioned using gold evidence locations: 40 pages for MMLongBench-Doc and 30 for LongDocURL.

\paragraph{Answer quality.}
We report generalized F1 on a 0--100 scale \citep{ma2024mmlongbench}. For each system, the same Claude Opus 4.8 model and extraction prompt convert the generated response into a short answer. A rule-based scorer compares this answer with the reference according to its answer format. Generalized F1 aggregates these scores while accounting for whether the system answers or abstains. Evaluation subsets used for paired comparisons are specified in Appendix~\ref{app:supp}.

\paragraph{Retrieval and cost.}
We measure evidence-page recall as the fraction of annotated evidence pages covered by retrieved content. For passage-based retrieval, passage precision is the fraction of retrieved passages attributed to annotated evidence pages. We also report the number of pages retrieved per question. For Chunk-RAG and RAPTOR, the passage budget limits the estimated total number of tokens in the retrieved text supplied to the model. We measure query cost using input tokens summed across model calls, including the initial outline and subsequent routing and reading rounds.

%% file: tables/datasets.tex
\begin{table}[t]
\centering\small
\setlength{\tabcolsep}{3pt}
\begin{tabular}{lrr}
\toprule
& MMLongBench & LongDocURL \\
\midrule
Questions                & 1{,}091 & 2{,}325 \\
Documents                & 135 & 396 \\
Pages, median / max      & 28 / 468 & 78 / 149 \\
No text layer            & \textbf{23\%} & 0\% \\
Evidence pages/question  & 1.88 & 1.62 \\
Non-text evidence        & 50.7\% & 44.5\% \\
\bottomrule
\end{tabular}
\caption{Dataset versions used in our experiments. Text availability is measured from our extraction output. Non-text evidence includes tables, figures, and layout.}
\label{tab:datasets}
\end{table}

%% file: sections/en/05_results.tex
\section{Results and Analysis}
\label{sec:results}

\subsection{Overall performance}
\label{sec:results:main}
\label{sec:results:docagent}
\input{tables/main_results}

Table~\ref{tab:main} compares VistaTree with retrieval baselines and evidence-conditioned references. VistaTree with flat selection achieves 43.5 F1 on MMLongBench-Doc and 43.2 on LongDocURL, outperforming DocAgent on both benchmarks and Chunk-RAG and RAPTOR on LongDocURL. Hierarchical navigation scores 42.1 and 38.5, respectively.

The gold-page oracle provides an empirical upper-bound reference for the local model under ideal evidence selection. It achieves 61.0 F1 on answerable MMLongBench-Doc questions with valid evidence-page annotations and 58.1 on LongDocURL. The window reference achieves 50.4 F1 on MMLongBench-Doc and 52.2 on LongDocURL.

We also examine whether a stronger model can reuse the locally constructed trees to reduce input-token usage. With Claude, VistaTree achieves 56.8 F1 on MMLongBench-Doc and 45.6 on LongDocURL while using fewer input tokens than full-document input on MMLongBench-Doc and evidence-window input on LongDocURL (Appendix Table~\ref{tab:claude_tokens}). This reduction in input-token usage is accompanied by lower answer quality than the corresponding full-document and evidence-window references.

\begin{figure*}[t]
\centering
\includegraphics[width=\textwidth]{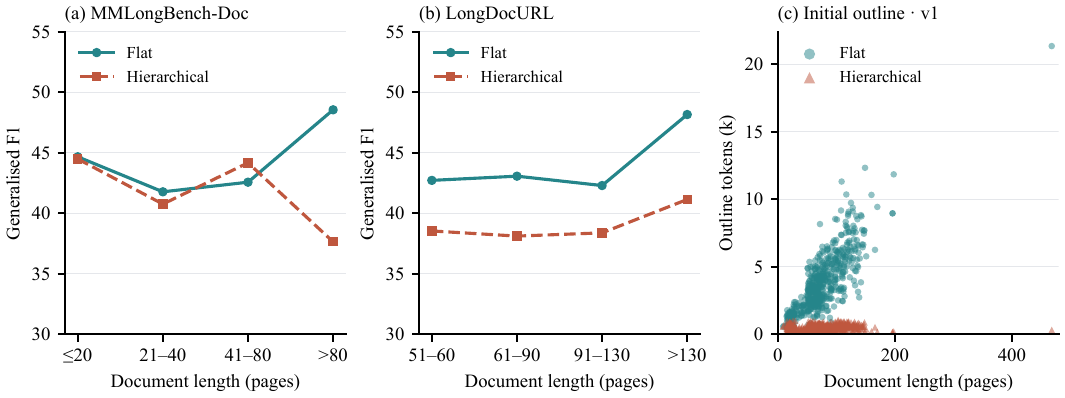}
\caption{Length and context cost for v1 trees. (a,b) Generalized F1 computed within document-length bins using the same questions with scored outputs from both navigation modes. Bins reflect the different document-length ranges and contain different documents and questions. (c) Initial-outline tokens measured with the Qwen3-VL tokenizer, with one point per document and navigation mode across both benchmarks.}
\label{fig:length}
\end{figure*}

\subsection{Navigation and summary rewriting}
\label{sec:results:navigation}
\input{tables/ablation}

We compare flat selection and hierarchical navigation on both benchmarks. With rewritten summaries, hierarchical navigation scores 1.4 F1 points lower on MMLongBench-Doc and 4.7 points lower on LongDocURL. Flat selection also achieves higher F1 with the original summaries (Table~\ref{tab:ablation}).

Summary rewriting improves F1 for both navigation modes on both benchmarks. Because the tree structure and page ranges remain fixed, the gains suggest that refining node descriptions can improve answer quality without modifying the underlying index structure.

\subsection{Document length and context cost}
\label{sec:results:length}
\label{sec:results:scaling}

Flat selection performs well across the evaluated document-length bins, achieving higher F1 than hierarchical navigation in all LongDocURL bins and most MMLongBench-Doc bins (Figure~\ref{fig:length}a,b). However, the flat outline grows with document length (Figure~\ref{fig:length}c), whereas hierarchical navigation offers a smaller initial outline. This difference reflects the trade-off between global summary visibility and progressive access, with the latter requiring additional navigation steps to reach relevant page spans.

\begin{figure*}[t]
\centering
\includegraphics[width=\textwidth]{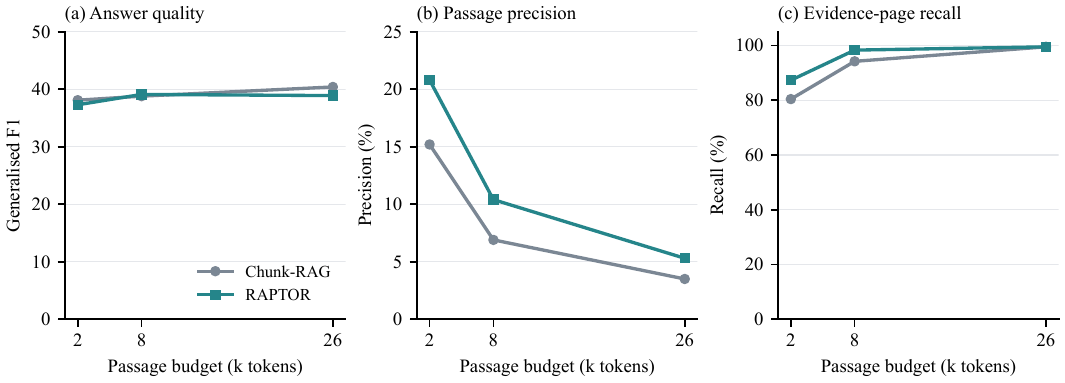}
\caption{Effect of the passage budget on LongDocURL. F1 is measured over the aggregate QA evaluation. Passage precision and evidence-page recall are measured on a 30-document sample. Points denote tested settings and lines connect them.}
\label{fig:budget}
\end{figure*}

Total query input also includes the subsequent navigation and page-reading calls. On the same v1 evaluation questions used for the length-bin comparison, hierarchical navigation uses 25,842 input tokens per question on MMLongBench-Doc, compared with 24,018 for flat selection. On LongDocURL, it uses fewer tokens: 29,334 versus 31,992. The overall query cost therefore depends on document length, which affects the initial outline size, as well as the subsequent navigation and reading calls.

\subsection{Retrieval budget and evidence relevance}
\label{sec:results:sweep}
\label{sec:results:recall}

We vary the passage budget to examine whether retrieving more content ranked by similarity provides better evidence for answering the question (Figure~\ref{fig:budget}). Increasing Chunk-RAG's budget from 2k to 26k raises F1 from 38.1 to 40.4. RAPTOR peaks at 39.1 with an 8k budget and scores 38.9 at 26k. Answer quality therefore changes only modestly despite the much larger retrieved context.

The retrieval diagnostics help explain this limited return. In the 30-document sample, both methods reach 99.5\% evidence-page recall at the largest budget, while Chunk-RAG's passage precision falls from 15.2\% to 3.5\% and RAPTOR's from 20.8\% to 5.3\%. Higher budgets thus cover more evidence pages while substantially expanding the context drawn from other pages. Moreover, page-level recall credits passages that share an annotated evidence page even when they do not contain the answer-bearing information.

These results highlight a limitation of similarity-based retrieval: passages that resemble the question may provide little evidence for answering it. Broader evidence-page coverage therefore does not necessarily translate into more useful context for question answering.

%% file: tables/main_results.tex
\begin{table}[!t]
\centering\small
\setlength{\tabcolsep}{3.5pt}
\begin{tabular}{lrr}
\toprule
Method & MMLongBench & LongDocURL \\
\midrule
Chunk-RAG & --- & 40.4 \\
RAPTOR & --- & 38.9 \\
DocAgent & 31.4 & 37.9 \\
\midrule
VistaTree, flat & \textbf{43.5} & \textbf{43.2} \\
VistaTree, hierarchical & 42.1 & 38.5 \\
\midrule
\multicolumn{3}{l}{\textit{Gold-evidence references (Qwen3-VL-8B-Instruct)}} \\
Oracle, gold pages$^\dagger$ & 61.0 & 58.1 \\
Evidence window$^\dagger$ & 50.4 & 52.2 \\
\midrule
\multicolumn{3}{l}{\textit{Commercial model, local index}} \\
VistaTree, Claude & 56.8 & 45.6 \\
\midrule
\multicolumn{3}{l}{\textit{Published human references}} \\
Human experts & 66.0 & 84.8$^\ddagger$ \\
\bottomrule
\end{tabular}
\caption{Question-answering F1, with published human references. The main comparison uses Qwen3-VL-8B-Instruct and v1 summaries for VistaTree. $^\dagger$Uses gold evidence locations (40-page windows for MMLongBench-Doc, 30 for LongDocURL). Claude searches an index built by the local model. Chunk-RAG and RAPTOR are reported on LongDocURL only, where extracted text is available for all documents. Their MMLongBench-Doc results are omitted because some documents lack extractable text under our preprocessing pipeline. Both use a 26k passage budget. A dash denotes an unreported configuration. Human scores are from \citet{ma2024mmlongbench} and \citet{longdocurl}. $^\ddagger$LongDocURL reports generalized accuracy on 102 sampled questions. Scoring subsets are defined in Appendix~\ref{app:supp}.}
\label{tab:main}
\end{table}

%% file: tables/ablation.tex
\begin{table}[!t]
\centering\small
\setlength{\tabcolsep}{3pt}
\begin{tabular}{llrr}
\toprule
Navigation & Tree & MMLongBench & LongDocURL \\
\midrule
Flat & v0 & 43.4 & 41.9 \\
Flat & v1 & \textbf{43.5} & \textbf{43.2} \\
\midrule
Hierarchical & v0 & 40.0 & 38.0 \\
Hierarchical & v1 & \textbf{42.1} & \textbf{38.5} \\
\bottomrule
\end{tabular}
\caption{Ablation of navigation and summary rewriting (F1). v0 denotes the original summaries and v1 the rewritten summaries. Bold marks the higher score within each navigation pair.}
\label{tab:ablation}
\end{table}

%% file: sections/en/06_conclusion.tex
\section{Discussion}
\label{sec:discussion}

Hierarchical navigation is intended for longer documents whose complete outlines may exceed the context budget. Our benchmark comparisons measure the gap relative to flat selection, which gives the agent access to all leaf summaries at once. Hierarchical navigation achieves lower aggregate F1 while requiring a smaller initial outline. However, the additional navigation and reading calls mean that this smaller outline does not consistently reduce total query input.

This trade-off is particularly relevant to long confidential documents that must be processed locally. Clinical study reports, for example, can approach or exceed a thousand pages. Our public-benchmark evaluation, however, includes documents only up to 468 pages. Evaluation at clinical-report scale is needed to establish whether the tree remains useful for evidence selection and how navigation affects answer quality and context usage as document length increases.

\section{Conclusion}
\label{sec:conclusion}
We introduced VistaTree, a document retrieval agent that uses a locally hosted vision--language model to construct and search summary trees from page images. Experiments on two public benchmarks demonstrate the effectiveness of VistaTree and quantify the gap between flat selection and hierarchical navigation, which is intended for longer documents whose complete outlines may exceed the context budget. Summary rewriting further improves performance while preserving the tree structure and page ranges. This approach supports document question answering in settings where confidentiality restricts external processing.

%% file: sections/en/07_limitations.tex
\paragraph{Document length and evaluation scope.}
Our evaluation uses two public benchmarks of English-language documents and a single local model. The longest documents contain 468 pages in MMLongBench-Doc and 149 in LongDocURL. Evaluation on substantially longer documents, including clinical study reports approaching a thousand pages, remains future work, as does evaluation in other languages and with different local models.

\paragraph{Index construction and query cost.}
Tree construction incurs an upfront inference cost for segmentation and summarization, with summary rewriting requiring an additional pass. Although the index can be reused, documents receiving few queries offer less opportunity to amortize this cost. Each query also incurs navigation and reading costs that depend on the search process.

\paragraph{Instruction following.}
The 8B model can struggle to follow multi-step navigation instructions and maintain the required tool-call format. Reliable execution depends on carefully designed prompts that specify the available actions, their arguments, and the navigation procedure. This dependence may require further prompt adaptation when changing the model or tool interface.

%% file: sections/en/08_appendix.tex
\section{Implementation Details}
\label{app:impl}

\paragraph{Tree construction.}
The tree builder operates on rendered page images. It processes one page per segmentation call, carrying a short running description of the current span. A forced split prevents a span from continuing beyond the configured accumulation limit, set to 30 pages by default. If the model identifies a topic boundary within the preceding page, that page can be shared by adjacent leaves.

Leaf summarization uses at most ten images, selected across the span, together with the running description. Higher levels group consecutive child nodes from their titles and shortened summaries, then generate a description of each group. Lists exceeding 80 nodes use windowed grouping. Construction aims for fewer than ten top-level nodes and includes a depth safety limit. Tree construction uses the 8B model with a 65,536-token server context and a twelve-image server limit.

\paragraph{Summary rewriting.}
Each node is rewritten against a frozen snapshot of its siblings' descriptions. Leaf nodes use up to ten sampled page images, while internal nodes use their children's original summaries. The implementation permits up to three sibling inspections per node and retains the original text if rewriting fails. In the reported MMLongBench-Doc build, 3,773 of 3,780 nodes were rewritten, seven retained their original text, and no sibling inspections were requested. Tree shape, node identifiers, and page ranges were unchanged. Mean summary length increased from 234 to 352 characters.

\paragraph{Navigation and history.}
The agent allows up to ten tool rounds before requesting a final answer. Flat selection exposes all leaves at the outset, whereas hierarchical navigation starts with the top-level nodes. The implementation retains recent interactions, removes older images as needed, and compresses older tool outputs when the estimated history budget is exceeded. Earlier page images are removed to conserve context, while the agent's textual observations and reasoning remain in the conversation, subject to history compression. The default limit on images retained in the conversation is 32. The query-serving deployment permits up to 40 images per request.

\paragraph{Reviewer.}
The optional reviewer receives the question, proposed answer, and pages viewed by the VistaTree agent. It labels the answer as supported, contradicted, or absent from the supplied pages. An unsupported answer is returned to the agent with feedback for reconsideration. The review configurations allow one such review round.

\section{Baseline Configurations}
\label{app:baselines}

\paragraph{Chunk-RAG and RAPTOR.}
Text passages target approximately 100 tokens, with sentence boundaries preserved where possible. Chunks end at page boundaries, so passages can be shorter than the target length. RAPTOR uses UMAP and Gaussian-mixture soft clustering, selects the number of mixture components by BIC, summarizes clusters recursively, and retrieves across the collapsed tree. Both use multilingual E5 embeddings \citep{e5instruct}, and RAPTOR uses the 8B model as its summarizer. Chunk sizes and passage budgets are estimated from text length using four characters per token. We evaluate both baselines under multiple passage budgets.

\paragraph{DocAgent.}
We use the official DocAgent implementation with the locally hosted Qwen3-VL-8B-Instruct model for question answering and the same answer extractor used for the other systems. Table~\ref{tab:additional} separately reports the published DocAgent result with Claude 3.5 Sonnet.

\section{Supplementary Results and Evaluation Details}
\label{app:supp}

\subsection{Additional model and retrieval configurations}
\input{tables/additional_configurations}

Table~\ref{tab:additional} reports additional model and retrieval configurations, including the published DocAgent reference and answer-review experiments with an earlier navigation loop.

\paragraph{Claude input-token usage.}
Table~\ref{tab:claude_tokens} reports Claude query input tokens summed across model calls. Input totals include uncached tokens, cache-creation tokens, and cache-read tokens, and exclude local index construction. MMLongBench-Doc values are the recorded run-level means, while LongDocURL values are recomputed on questions with scored outputs from both configurations. These figures measure token volume rather than billed API cost.

\begin{table}[t]
\centering\small
\setlength{\tabcolsep}{2pt}
\begin{tabular}{lrr}
\toprule
Configuration & MMLongBench & LongDocURL \\
\midrule
Claude, full PDF & 109{,}130 & --- \\
Claude, 30-page window & --- & 75{,}056 \\
Claude, VistaTree flat & 28{,}565 & 26{,}110 \\
\midrule
Input-token reduction & 73.8\% & 65.2\% \\
\bottomrule
\end{tabular}
\caption{Mean Claude query input tokens per question. Reduction is relative to full-PDF input on MMLongBench-Doc and the evidence-conditioned 30-page window on LongDocURL. VistaTree uses locally constructed v0 trees.}
\label{tab:claude_tokens}
\end{table}

\subsection{Evaluation coverage and scoring}

\paragraph{MMLongBench-Doc.}
The benchmark contains 1,091 questions. The gold-page oracle is evaluated on 837 questions, excluding 246 annotated as unanswerable and eight with invalid evidence-page references. The full-PDF Claude run, the 40-page window, and hierarchical navigation with review each have 1,085 scored questions. The other reported configurations have 1,091. The 40-page window excludes six questions from a document whose window exceeds the context limit. The published DocAgent reference uses 1,082 questions.

\paragraph{LongDocURL.}
The benchmark contains 2,325 questions. Claude with an evidence-conditioned window has 2,295 scored questions, the gold-page oracle 2,316, and the local model with an evidence-conditioned window 2,252. The other reported configurations have 2,318 scored questions. The v1 flat and hierarchical runs, Chunk-RAG at 26k, RAPTOR at 26k, and DocAgent are compared on their 2,318 jointly completed questions. DocAgent input-token usage is unavailable, so its comparison is limited to answer quality.

\paragraph{Evidence-conditioned reading.}
As a supplementary comparison, we recompute scores for the oracle, the 40-page window, and flat selection on their 832 jointly completed MMLongBench-Doc questions. They achieve 60.9, 52.9, and 45.6 F1, respectively. This comparison uses the same questions, with different supplied context and interaction procedures.

\subsection{Published human references}
The MMLongBench-Doc paper reports human experts at 66.0 generalized F1 (65.8 generalized accuracy) on its original benchmark version \citep{ma2024mmlongbench}. LongDocURL reports 84.8 generalized accuracy on 102 uniformly distributed samples \citep{longdocurl}. Table~\ref{tab:main} retains each paper's reported metric and evaluation scope.

\subsection{Supplementary retrieval diagnostics}
\input{tables/navigation}
\label{sec:results:nav}
\paragraph{Answer review.}
The reviewer experiments use the original summaries and an earlier navigation loop. On their respective shared questions, review changes flat F1 from 45.3 to 45.6 and hierarchical F1 from 39.7 to 42.0.

Table~\ref{tab:nav} characterizes hierarchical navigation on LongDocURL using the earlier v0 configuration. Opened nodes contain gold evidence 48.9\% of the time versus 17.3\% for random siblings, while 8.0\% of questions reach the round limit.

\paragraph{Reading volume.}
The reported flat configuration reads approximately 4.4 complete pages per question. This measure counts page images rather than the text passages used by Chunk-RAG and RAPTOR.

\subsection{Document-length analysis}
The length-bin comparisons use questions with scored outputs from both navigation modes. Valid abstentions are included in generalized F1, while questions with failed runs in either mode are excluded rather than assigned zero scores.

In increasing order of document length, the v1 MMLongBench-Doc groups in Figure~\ref{fig:length} contain 265, 430, 259, and 137 shared questions over 32, 55, 31, and 17 documents, respectively. The LongDocURL groups contain 342, 968, 800, and 208 shared questions over 71, 177, 117, and 30 documents. For each bin, generalized F1 is recomputed from the answerability flags and score sums. Outline counts use \texttt{Qwen/Qwen3-VL-8B-Instruct} with special tokens disabled, for both navigation modes and all 531 documents.

%% file: tables/additional_configurations.tex
\begin{table}[t]
\centering\small
\setlength{\tabcolsep}{4pt}
\begin{tabular}{lrr}
\toprule
Configuration & MMLongBench & LongDocURL \\
\midrule
Claude, full PDF & 67.6 & --- \\
Claude, window$^\dagger$ & --- & 60.9 \\
RAPTOR, 8k budget & --- & 39.1 \\
DocAgent, published$^*$ & 54.1 & --- \\
\midrule
\multicolumn{3}{l}{\textit{Earlier v0 navigation}} \\
Flat & 45.3 & --- \\
Flat + review & 45.6 & --- \\
Hierarchical & 39.7 & --- \\
Hierarchical + review & 42.0 & --- \\
\bottomrule
\end{tabular}
\caption{Additional configurations (F1). $^\dagger$Uses gold evidence locations. $^*$Published DocAgent result with Claude 3.5 Sonnet \citep{sun2025docagent}, using a different evaluation setup. Earlier v0 navigation rows report review runs and their paired baselines.}
\label{tab:additional}
\end{table}

%% file: tables/navigation.tex
\begin{table}[!htbp]
\centering\small
\begin{tabular}{lr}
\toprule
Nodes opened per question & 2.85 (1.40 leaves) \\
Mean descent depth & 1.93 of 3 \\
Opened node holds gold evidence & \textbf{48.9\%} \\
\quad random sibling & 17.3\% \\
\texttt{go\_back} used & 20.5\%, 0 failures \\
Nodes re-opened & 0\% \\
Reached the 10-round cap & 8.0\% \\
\bottomrule
\end{tabular}
\caption{Earlier v0 hierarchical navigation on LongDocURL. The opened-node evidence rate is compared with random selection among sibling nodes.}
\label{tab:nav}
\end{table}

%% file: refs.bib
@inproceedings{lewis2020rag,
  title = {Retrieval-Augmented Generation for Knowledge-Intensive {NLP} Tasks},
  author = {Lewis, Patrick and Perez, Ethan and Piktus, Aleksandra and Petroni, Fabio and Karpukhin, Vladimir and Goyal, Naman and K{\"u}ttler, Heinrich and Lewis, Mike and Yih, Wen-tau and Rockt{\"a}schel, Tim and Riedel, Sebastian and Kiela, Douwe},
  booktitle = {Advances in Neural Information Processing Systems},
  volume = {33},
  year = {2020},
  url = {https://arxiv.org/abs/2005.11401}
}

@inproceedings{karpukhin2020dpr,
  title = {Dense Passage Retrieval for Open-Domain Question Answering},
  author = {Karpukhin, Vladimir and Oguz, Barlas and Min, Sewon and Lewis, Patrick and Wu, Ledell and Edunov, Sergey and Chen, Danqi and Yih, Wen-tau},
  booktitle = {Proceedings of the 2020 Conference on Empirical Methods in Natural Language Processing},
  year = {2020},
  pages = {6769--6781},
  doi = {10.18653/v1/2020.emnlp-main.550},
  url = {https://aclanthology.org/2020.emnlp-main.550/}
}

@article{liu2024lost,
  title = {Lost in the Middle: How Language Models Use Long Contexts},
  author = {Liu, Nelson F. and Lin, Kevin and Hewitt, John and Paranjape, Ashwin and Bevilacqua, Michele and Petroni, Fabio and Liang, Percy},
  journal = {Transactions of the Association for Computational Linguistics},
  volume = {12},
  pages = {157--173},
  year = {2024},
  doi = {10.1162/tacl_a_00638},
  url = {https://aclanthology.org/2024.tacl-1.9/}
}

@inproceedings{kim2022donut,
  title = {{OCR}-free Document Understanding Transformer},
  author = {Kim, Geewook and Hong, Teakgyu and Yim, Moonbin and Nam, Jeongyeon and Park, Jinyoung and Yim, Jinyeong and Hwang, Wonseok and Yun, Sangdoo and Han, Dongyoon and Park, Seunghyun},
  booktitle = {European Conference on Computer Vision},
  year = {2022},
  url = {https://arxiv.org/abs/2111.15664}
}

@inproceedings{lee2023pix2struct,
  title = {{Pix2Struct}: Screenshot Parsing as Pretraining for Visual Language Understanding},
  author = {Lee, Kenton and Joshi, Mandar and Turc, Iulia and Hu, Hexiang and Liu, Fangyu and Eisenschlos, Julian and Khandelwal, Urvashi and Shaw, Peter and Chang, Ming-Wei and Toutanova, Kristina},
  booktitle = {Proceedings of the 40th International Conference on Machine Learning},
  year = {2023},
  url = {https://arxiv.org/abs/2210.03347}
}

@inproceedings{sarthi2024raptor,
  title = {{RAPTOR}: Recursive Abstractive Processing for Tree-Organized Retrieval},
  author = {Sarthi, Parth and Abdullah, Salman and Tuli, Aditi and Khanna, Shubh and Goldie, Anna and Manning, Christopher D.},
  booktitle = {International Conference on Learning Representations},
  year = {2024},
  url = {https://arxiv.org/abs/2401.18059}
}

@inproceedings{sun2025docagent,
  title = {{DocAgent}: An Agentic Framework for Multi-Modal Long-Context Document Understanding},
  author = {Sun, Li and He, Liu and Jia, Shuyue and He, Yangfan and You, Chenyu},
  booktitle = {Proceedings of the 2025 Conference on Empirical Methods in Natural Language Processing},
  pages = {17701--17716},
  year = {2025},
  doi = {10.18653/v1/2025.emnlp-main.893},
  url = {https://aclanthology.org/2025.emnlp-main.893/}
}

@inproceedings{ma2024mmlongbench,
  title = {{MMLongBench-Doc}: Benchmarking Long-context Document Understanding with Visualizations},
  author = {Ma, Yubo and Zang, Yuhang and Chen, Liangyu and Chen, Meiqi and Jiao, Yizhu and Li, Xinze and Lu, Xinyuan and Liu, Ziyu and Ma, Yan and Dong, Xiaoyi and Zhang, Pan and Pan, Liangming and Jiang, Yu-Gang and Wang, Jiaqi and Cao, Yixin and Sun, Aixin},
  booktitle = {Advances in Neural Information Processing Systems, Datasets and Benchmarks Track},
  year = {2024},
  url = {https://arxiv.org/abs/2407.01523}
}

@inproceedings{longdocurl,
  title = {{LongDocURL}: a Comprehensive Multimodal Long Document Benchmark Integrating Understanding, Reasoning, and Locating},
  author = {Deng, Chao and Yuan, Jiale and Bu, Pi and Wang, Peijie and Li, Zhong-Zhi and Xu, Jian and Li, Xiao-Hui and Gao, Yuan and Song, Jun and Zheng, Bo and Liu, Cheng-Lin},
  booktitle = {Proceedings of the 63rd Annual Meeting of the Association for Computational Linguistics (Volume 1: Long Papers)},
  pages = {1135--1159},
  year = {2025},
  doi = {10.18653/v1/2025.acl-long.57},
  url = {https://aclanthology.org/2025.acl-long.57/}
}

@misc{pageindex,
  title = {{PageIndex}: Next-Generation Vectorless, Reasoning-based {RAG}},
  author = {Zhang, Mingtian and Tang, Yu},
  year = {2025},
  howpublished = {PageIndex Blog},
  url = {https://pageindex.ai/blog/pageindex-intro},
  note = {Published 19 September 2025}
}

@article{qwen3vl,
  title = {{Qwen3-VL} Technical Report},
  author = {Shuai Bai and Yuxuan Cai and Ruizhe Chen and Keqin Chen and Xionghui Chen and Zesen Cheng and Lianghao Deng and Wei Ding and Chang Gao and Chunjiang Ge and Wenbin Ge and Zhifang Guo and Qidong Huang and Jie Huang and Fei Huang and Binyuan Hui and Shutong Jiang and Zhaohai Li and Mingsheng Li and Mei Li and Kaixin Li and Zicheng Lin and Junyang Lin and Xuejing Liu and Jiawei Liu and Chenglong Liu and Yang Liu and Dayiheng Liu and Shixuan Liu and Dunjie Lu and Ruilin Luo and Chenxu Lv and Rui Men and Lingchen Meng and Xuancheng Ren and Xingzhang Ren and Sibo Song and Yuchong Sun and Jun Tang and Jianhong Tu and Jianqiang Wan and Peng Wang and Pengfei Wang and Qiuyue Wang and Yuxuan Wang and Tianbao Xie and Yiheng Xu and Haiyang Xu and Jin Xu and Zhibo Yang and Mingkun Yang and Jianxin Yang and An Yang and Bowen Yu and Fei Zhang and Hang Zhang and Xi Zhang and Bo Zheng and Humen Zhong and Jingren Zhou and Fan Zhou and Jing Zhou and Yuanzhi Zhu and Ke Zhu},
  journal = {arXiv preprint arXiv:2511.21631},
  year = {2025},
  url = {https://arxiv.org/abs/2511.21631}
}

@article{e5instruct,
  title = {Multilingual {E5} Text Embeddings: A Technical Report},
  author = {Wang, Liang and Yang, Nan and Huang, Xiaolong and Yang, Linjun and Majumder, Rangan and Wei, Furu},
  journal = {arXiv preprint arXiv:2402.05672},
  year = {2024},
  url = {https://arxiv.org/abs/2402.05672}
}

@inproceedings{weller2025followir,
  title = {{FollowIR}: Evaluating and Teaching Information Retrieval Models to Follow Instructions},
  author = {Weller, Orion and Chang, Benjamin and MacAvaney, Sean and Lo, Kyle and Cohan, Arman and Van Durme, Benjamin and Lawrie, Dawn and Soldaini, Luca},
  booktitle = {Proceedings of the 2025 Conference of the Nations of the Americas Chapter of the Association for Computational Linguistics: Human Language Technologies (Volume 1: Long Papers)},
  pages = {11926--11942},
  year = {2025},
  doi = {10.18653/v1/2025.naacl-long.597},
  url = {https://aclanthology.org/2025.naacl-long.597/}
}

@inproceedings{
jin2025longcontext,
title={Long-Context {LLM}s Meet {RAG}: Overcoming Challenges for Long Inputs in {RAG}},
author={Bowen Jin and Jinsung Yoon and Jiawei Han and Sercan O Arik},
booktitle={The Thirteenth International Conference on Learning Representations},
year={2025},
url={https://openreview.net/forum?id=oU3tpaR8fm}
}

@inproceedings{saadfalcon2024pdftriage,
  title = {{PDFTriage}: Question Answering over Long, Structured Documents},
  author = {Saad-Falcon, Jon and Barrow, Joe and Siu, Alexa and Nenkova, Ani and Yoon, Seunghyun and Rossi, Ryan A. and Dernoncourt, Franck},
  booktitle = {Proceedings of the 2024 Conference on Empirical Methods in Natural Language Processing: Industry Track},
  pages = {153--169},
  year = {2024},
  doi = {10.18653/v1/2024.emnlp-industry.13},
  url = {https://aclanthology.org/2024.emnlp-industry.13/}
}

@article{poznanski2025olmocr,
  title={{olmOCR}: Unlocking Trillions of Tokens in {PDFs} with Vision Language Models},
  author={Poznanski, Jake and Rangapur, Aman and Borchardt, Jon and Dunkelberger, Jason and Huff, Regan and Lin, Daniel and Wilhelm, Christopher and Lo, Kyle and Soldaini, Luca},
  journal={arXiv preprint arXiv:2502.18443},
  year={2025}
}

@article{poznanski2025olmocr2,
  title={{olmOCR} 2: Unit Test Rewards for Document {OCR}},
  author={Poznanski, Jake and Soldaini, Luca and Lo, Kyle},
  journal={arXiv preprint arXiv:2510.19817},
  year={2025}
}

@inproceedings{poznanski2026olmocr,
  title={The {olmOCR} Project: Building Fully Open {OCR} using {VLMs}},
  author={Poznanski, Jake and Lo, Kyle and Soldaini, Luca},
  booktitle={Proceedings of the 64th Annual Meeting of the Association for Computational Linguistics (Volume 3: System Demonstrations)},
  pages={626--635},
  year={2026}
}

@article{auer2024docling,
  title={{Docling} Technical Report},
  author={Auer, Christoph and Lysak, Maksym and Nassar, Ahmed and Dolfi, Michele and Livathinos, Nikolaos and Vagenas, Panos and Ramis, Cesar Berrospi and Omenetti, Matteo and Lindlbauer, Fabian and Dinkla, Kasper and others},
  journal={arXiv preprint arXiv:2408.09869},
  year={2024}
}

@article{livathinos2025docling,
  title={{Docling}: An Efficient Open-Source Toolkit for {AI}-driven Document Conversion},
  author={Livathinos, Nikolaos and Auer, Christoph and Lysak, Maksym and Nassar, Ahmed and Dolfi, Michele and Vagenas, Panos and Ramis, Cesar Berrospi and Omenetti, Matteo and Dinkla, Kasper and Kim, Yusik and others},
  journal={arXiv preprint arXiv:2501.17887},
  year={2025}
}

@article{wang2024mineru,
  title={{MinerU}: An Open-Source Solution for Precise Document Content Extraction},
  author={Wang, Bin and Xu, Chao and Zhao, Xiaomeng and Ouyang, Linke and Wu, Fan and Zhao, Zhiyuan and Xu, Rui and Liu, Kaiwen and Qu, Yuan and Shang, Fukai and others},
  journal={arXiv preprint arXiv:2409.18839},
  year={2024}
}

@inproceedings{blecher2024nougat,
  title={{Nougat}: Neural Optical Understanding for Academic Documents},
  author={Blecher, Lukas and Cucurull Preixens, Guillem and Scialom, Thomas and Stojnic, Robert},
  booktitle={International Conference on Learning Representations},
  volume={2024},
  pages={37646--37663},
  year={2024}
}
